\documentclass{article} % For LaTeX2e
\usepackage[numbers]{natbib}
\usepackage[accepted]{icml2021}
\usepackage{amsmath}
\usepackage[utf8]{inputenc} % allow utf-8 input
\usepackage[T1]{fontenc}    % use 8-bit T1 fonts
\usepackage{hyperref}       % hyperlinks
\usepackage{booktabs}       % professional-quality tables
\usepackage{amsfonts}       % blackboard math symbols
\usepackage{microtype}
\usepackage{graphicx}
\usepackage{tikz}
\usetikzlibrary{bayesnet}
\usepackage{subcaption}
\begin{document}

\twocolumn[
\icmltitle{
Invariance-based Multi-Clustering of Latent Space Embeddings for Equivariant Learning}

\begin{icmlauthorlist}
\icmlauthor{Chandrajit Bajaj}{CS}
\icmlauthor{Avik Roy}{Phys}
\icmlauthor{Haoran Zhang}{CS}
\end{icmlauthorlist}

\icmlaffiliation{CS}{Department of Computer Science, The University of Texas at Austin, Austin, TX 78712}
\icmlaffiliation{Phys}{Department of Physics, The University of Texas at Austin, Austin, TX 78712}

\icmlcorrespondingauthor{Chandrajit Bajaj}{bajaj@cs.utexas.edu}
\icmlcorrespondingauthor{Avik Roy}{aroy@utexas.edu}
\icmlcorrespondingauthor{Haoran Zhang}{hrzhang@cs.utexas.edu}

\vskip 0.3in
]
% \author{
%     Chandrajit Bajaj\\
    
%     \texttt{bajaj@cs.utexas.edu}
%     %\And
%     %Grant Kluber\\
%   % Department of Computer Science\\
%     %University of Texas at Austin\\
%   % Austin, TX 78712 \\
%   % \texttt{gkluber@utexas.edu}
%     \And
%     Avik Roy \\
%     Department of Physics\\
%     University of Texas at Austin\\
%     Austin, TX 78712 \\
%     \texttt{aroy@utexas.edu}
%     \And
%     Haoran Zhang\\
%     Department of Computer Science\\
%     University of Texas at Austin\\
%     Austin, TX 78712 \\
%     \texttt{hrzhang@cs.utexas.edu}
%     }

%\maketitle 
\printAffiliationsAndNotice{}
\begin{abstract}
    Variational Autoencoders (VAEs) have been shown to be remarkably effective in recovering model latent spaces for several computer vision tasks. 
    However, currently trained VAEs, for a number of reasons, seem to fall short in learning invariant and equivariant clusters in latent space. 
    Our work focuses on providing solutions to this problem and presents an approach to disentangle equivariance feature maps in a Lie group manifold by enforcing deep, group-invariant learning.  
    Simultaneously implementing a novel separation of semantic and equivariant variables of the latent space representation, we formulate a modified Evidence Lower BOund (ELBO) by using a mixture model pdf like Gaussian mixtures for invariant cluster embeddings that allows superior unsupervised variational clustering.
    Our experiments show that this model effectively learns to disentangle the invariant and equivariant representations with significant improvements in the  learning rate and an observably superior image recognition and canonical state reconstruction compared to the currently best deep learning models.
\end{abstract}

\section{Introduction}
In multi-category visual data, a main problem of interest is object classification while simultaneously learning any invariant information that is present. Our data is often an ensemble of images where multiple members of the dataset are related with one another by simple  geometric transformations. Such transformations are often modeled by coordinate transformation matrices on a Lie manifold. We are interested in fleshing out a canonical reconstruction that is invariant under such group transformations while independently learning the transformation that faithfully rederives the equivariant reconstruction of image data \citep{Zhu:2019motion, Reisert:2007}. 
For these tasks, un-supervised or semi-supervised learning methods like variational auto-encoders (VAE) \cite{Kingma:2014, Tomczak:2018} have shown superior potential for learning meaningful latent representation of the image. 

However, vanilla VAE models easily confound invariant representation with non-equivariant variational perturbations.
Various papers have called attention to the inability of generative models like GAN \cite{Goodfellow:2014} and VAE \cite{Kingma:2014} in faithful factorization of latent representations \cite{Ridgeway:2016}. Notably, the unsupervised InfoGAN \cite{Chen:2016}, $\beta$-VAE \cite{Higgins:2017}, $\beta$-TCVAE \cite{Chen:2018}, factor-VAE \cite{Kim:2018},  and the semi-supervised DC-IGN \cite{Kulkarni:2015} are some examples of the necessary effort needed in learning factorized representations with modified generative models. These methods, however, aim to learn the latent representations without any constraints on the latent space. Moreover, InfoGan faces the problem of instability in training, while $\beta$-VAE suffers from reconstruction accuracy as a trade-off for attempting disentanglement \cite{Burgess:2018}. Some other methods like CSVAE \cite{Klys:2018} and ADGM \cite{Maale:2016} learn the representations with structured latent spaces, while mostly through supervised or semi-supervised training\cite{Scott:2014,cohen2016group,tai2019equivariant,fuchs2020se}. Recent works like $\beta$-TCVAE \cite{Chen:2018} and factor-VAE \cite{Kim:2018} based on the idea of $\beta$-VAE introduce total correlation and dimension-wise KL (Kullback-Leibler) divergence in the marginal KL expression, while others like AVT tackle this problem from an information-theoretic perspective by maximizing the mutual information between transformations and representations \cite{Qi:2019, Zhang:2019}.

In recent work, affine-VAE \cite{Bidart:2019} and spatial-VAE \cite{Bepler:2019}  proposed similar feed-forward frameworks for disentangling variational state variables such as shape, color, lighting,  from affine transformational representations. The affine-VAE network builds up on a pre-optimized VAE for canonical image representations to estimate an optimized affine transformation rendering it an essentially supervised model. Without the guarantee of equivariance, the learned invariant representation by affine-VAE and spatial-VAE are not stable under spatial transformations. This  will be further elaborated in detail in section $4$. 

A second prong to this problem is to obtain faithful classification in variational settings. Deep CNNs like H-nets \cite{Worrall:2017} and LieConv \cite{Finzi:2020} implement invariant image classification under rotation and translation. However, these approaches do not address classification in a variational setting. On the other hand, variational clustering architectures \cite{jiang:2017, dilokthanakul:2016, lim:2020} often don't consider invariant clustering, i.e. classifying the image content independently of the group transformations. In this paper, we present a hierarchical framework, Multi-Clustering Equivariant VAE (MCE-VAE), that simultaneously learns a multi-cluster invariant and equivariant latent representations of dynamic data. We present a simple version of MCE-VAE  with independent learning units for the variational, categorical, and transformational representations of objects. In section 2, we briefly explain the algebra of Lie manifolds for generic spatial transformations and an overview of spatial transformation network \cite{Jaderberg:2015}. In section 3, we lay out the architecture of MCE-VAE and formulate the objective function for both supervised and unsupervised learning settings.
Finally in section 4, we test the performance of MCE-VAE concerning reconstruction accuracy and clustering ability against state-of-the-art benchmarks using SO(2) and SE(2) transformed MNIST data.

\section{Background}

\begin{figure}
\centering
\scalebox{0.7}{
    \begin{tikzpicture}[pile/.style={thick, ->, >=stealth', shorten <=2pt, shorten
    >=2pt}, dpile/.style={thick, dashed, ->, >=stealth', shorten <=2pt, shorten
    >=2pt}, scale=.7]
    \path[draw, fill=gray!40] (0,0) -- (0,2) -- (2,2) -- (2,0) -- (0,0);
    \path[draw, fill=gray!40] (0.5,0.5) -- (0.5,2.5) -- (2.5,2.5) -- (2.5,0.5) -- (0.5,0.5);
    \draw (-1, -0.5) node[right] {input image};
    \draw[pile] (2.6, 1) -- (5.9,1) node[pos=0.5, above]{Transformation};
    \draw[pile] (2.6, 1) -- (5.9,1) node[pos=0.5, below]{Encoder};
    \path[draw, fill=gray!40] (6,0) -- (6,2) -- (8,2) -- (8,0) -- (6,0);
    \draw (8, 1.7) node[left] {$\mathbb{R}^n$};
    \draw (4.8, -0.5) node[right] {coefficient vector};
    \fill[red!80] (7.2, 0.4) circle (2pt);
    \draw (7.2, 0.4) node[right] {$\tau$};
    \draw[dpile] (8.1, 1) -- (10.9,-0.3);
    \draw (9.8, -0.2) node[left] {$H$};
    \path[draw, fill=gray!40] (10,-1) to[out=10,in=140] (13,-2) -- (15,-1) to[out=120,in=-10] (12,0.5) -- cycle;
    \draw (12, -1.5) node[left] {Lie Group $L$};
    \fill[red!80] (12, -0.2) circle (2pt);
    \draw (12, -0.2) node[right] {$M$};
    \draw[pile] (12.2, 0.5) -- (12.2,2);
    \draw (12.5, 1.2) node[right] {$\exp(\mathfrak{m})$};
    \draw[pile] (12.4,2) -- (12.4, 0.5) ;
    \draw (12.2, 1.2) node[left] {$\log(M)$};
    \draw[pile] (8.1, 1) -- (11,3);
    \draw (9.8, 2.5) node[left] {$G$};
    \path[draw, fill=gray!40] (11,2) -- (11,4) -- (13,4) -- (13,2) -- (11,2);
    \draw (10.5, 4.4) node[right] {Lie algebra $\mathfrak{l}$};
    \fill[red!80] (12, 3.1) circle (2pt);
    \draw (12, 3.1) node[right] {$\mathfrak{m}$};
    \end{tikzpicture}
    }
    \caption{The relationship between input data, coefficient vectors, Lie Group, and corresponding Lie algebra for 2D rigid transformations. The learned coefficient vector $\tau$ is mapped to the Lie algebra with a linear mapping with generator matrices $G$. The image of this map in the Lie algebra $\mathfrak{m}$ maps to the element $M$ in the Lie group, via matrix  exponentiation $exp(m)$ maps.}
    \label{fig:map}
\end{figure}
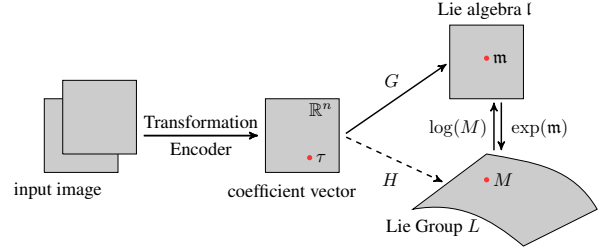

\paragraph{Lie algebra representation of spatial transformations.}\label{sec:Lie}
Lie groups are continuous topological groups that are also differentiable manifolds \cite{fulton:1991}. The tangent space at the identity element of this smooth manifold forms a vector space called the Lie algebra. Generally, the Lie algebra represents a vector space of infinitesimal transformations about the Lie group unity and spanned by a finite set of basis elements called the generators. For instance, the group of 2D rigid spatial transformations, SE(2) can be represented by transformation matrices \citep{Bluman:1989}, which are elements of corresponding Lie groups. The composition of these transformations correspond to the matrix multiplication in the Lie groups. The basis elements for this group's Lie Algebra, $\mathfrak{se}(2)$, represent infinitesimal planar rotation and translation in orthogonal directions that can be non-linearly compounded  to obtain finite transformations. The elements of Lie algebra  translate to the members of the Lie group via exponential mapping. If $m$ represents a linear combination of the generators $(G_i)$ of the Lie algebra $(\mathfrak{l})$, i.e. $m = \sum \tau_iG_i \in \mathfrak{l}$, then the corresponding member $(M)$ of the Lie group $L$ is obtained by matrix exponentiation, $M = \exp(m) \in L$ (Figure \ref{fig:map}).

\paragraph{Spatial transformer network.}\label{sec:stn}
Jaderberg et al. \cite{Jaderberg:2015} proposed the Spatial Transformer Network (STN) to deal with gradient propagating transformations of images. The STN consists of three parts: localization net, grid generator, and sampler. The localization net is a CNN that predicts the vectorized transformation matrix. For example, for 2D rigid transformation, the transformation matrix $M \in SE(2)$ follows
\begin{equation}
    M = \begin{pmatrix}
    \cos(\omega) & -\sin(\omega) & t_x\\
    \sin(\omega) & \cos(\omega) & t_y\\
    0 & 0 & 1
    \end{pmatrix}
\end{equation}
where $\omega, t_x, t_y$ respectively represents planar rotation and translation along the X and Y axes. The localization net predicts a vector $\pmb{\theta} = [\theta_1, \theta_2, \theta_3, \theta_4, \theta_5, \theta_6]$ such that
\begin{equation}\label{eqn:vec_trans}
    \begin{pmatrix}
    \theta_1 & \theta_2 & \theta_3\\
    \theta_4 & \theta_5 & \theta_6
    \end{pmatrix} = \begin{pmatrix}
    \cos(\omega) & -\sin(\omega) & t_x\\
    \sin(\omega) & \cos(\omega) & t_y\\
    \end{pmatrix}
\end{equation}

The grid generator maps the grid of each pixel in the input image to a transformed grid corresponding to the $\pmb{\theta}$ obtained from the localization net by
\begin{equation}
\begin{aligned}
\begin{pmatrix} x_{i}^{t} \\ y_{i}^{t} \end{pmatrix} 
= M\begin{pmatrix} x_{i}^{s} \\ y_{i}^{s} \\ 1\end{pmatrix}
= \begin{pmatrix}
    \theta_1 & \theta_2 & \theta_3\\
    \theta_4 & \theta_5 & \theta_6
    \end{pmatrix}\begin{pmatrix} x_{i}^{s} \\ y_{i}^{s} \\ 1\end{pmatrix}\\
\end{aligned}
\end{equation}
where $(x_i^s, y_i^s)$ is the inital coordinate of pixel $i$ and $(x_i^t, y_i^t)$ is the coordinate in the transformed grid.

The output of the first two steps gives transformed grid, but the mapping from the grid to the image is discrete and hardly differentiable, which hinders backpropagation in training. \citet{Jaderberg:2015} used a sampler to solve this problem. They define the output $V_i^c$ of pixel $i$ with transformed position $(x_i^t, y_i^t)$ and channel $c$ follows
\begin{equation}\label{eqn:stn}
    V_i^c = \sum_n\sum_m U_{n, m}^ck(x_i^s - m; \Phi_x)k(y_i^s - n; \Phi_y)
\end{equation}
where $U_{n, m}$ is the value at position $(m, n)$ in the input feature map, $k$ is some sampling kernel, and $\Phi$ is the parameter of $k$.

\section{MCE-VAE Architecture and Modified Evidence Lower Bound (ELBO) }\label{sec:method}

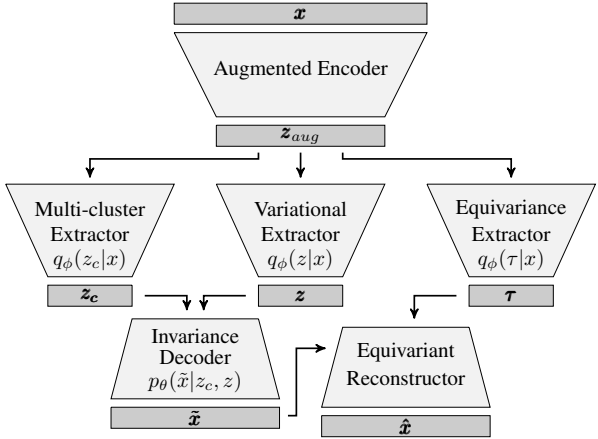
\begin{figure}
\centering
    \scalebox{0.8}{
    \begin{tikzpicture}
    [pile/.style={thick, ->, >=stealth', shorten <=2pt, shorten
    >=2pt}, dpile/.style={thick, dashed, ->, >=stealth', shorten <=2pt, shorten
    >=2pt}, scale=.7]
    \path[draw, fill=gray!40] (0,0.2) -- (6,0.2) -- (6,0.7) -- (0,0.7) -- (0,0.2);
    \draw (3, 0.45) node {$\pmb{x}$};
    
    \path[draw, fill=gray!10] (0,-0) -- (6,-0) -- (5,-2) -- (1,-2) -- (0,-0);
    \draw (3, -0.9) node {Augmented Encoder};
    
    \path[draw, fill=gray!40] (1,-2.2) -- (5,-2.2) -- (5,-2.7) -- (1,-2.7) -- (1, -2.2);
    \draw (3, -2.45) node {$\pmb{z}_{aug}$};
    
    \draw[pile] (2, -2.75) -- (2,-3) -- (-2, -3) -- (-2, -3.55); 
    \draw[pile] (4, -2.75) -- (4,-3) -- (8, -3) -- (8, -3.55);
    \draw[pile] (3, -2.75) -- (3, -3.55);
    
    \path[draw, fill=gray!10] (-4,-3.6) -- (0,-3.6) -- (-1,-5.8) -- (-3,-5.8) -- (-4,-3.6);
    \draw (-2, -4.2) node {Multi-cluster};
    \draw (-2, -4.8) node {Extractor};
    
    \draw (-2, -5.4) node {$q_\phi(z_c|x)$};
    
    \path[draw, fill=gray!40] (-1,-6.0) -- (-3,-6.0) -- (-3,-6.5) -- (-1,-6.5) -- (-1, -6.0);
    \draw (-2, -6.25) node {$\pmb{z_c}$};
    
    \path[draw, fill=gray!10] (1,-3.6) -- (5,-3.6) -- (4,-5.8) -- (2,-5.8) -- (1,-3.6);
    \draw (3, -4.2) node {Variational};
    \draw (3, -4.8) node {Extractor};
    
    \draw (3, -5.4) node {$q_\phi(z|x)$};
    
    \path[draw, fill=gray!40] (4,-6.0) -- (2,-6.0) -- (2,-6.5) -- (4,-6.5) -- (4, -6.0);
    \draw (3, -6.25) node {$\pmb{z}$};
    
    \path[draw, fill=gray!10] (6,-3.6) -- (10,-3.6) -- (9,-5.8) -- (7,-5.8) -- (6,-3.6);
    \draw (8, -4.2) node {Equivariance};
    \draw (8, -4.8) node {Extractor};
    
    \draw (8, -5.4) node {$q_\phi(\tau|x)$};
    
    \path[draw, fill=gray!40] (7,-6.0) -- (9,-6.0) -- (9,-6.5) -- (7,-6.5) -- (7, -6.0);
    \draw (8, -6.25) node {$\pmb{\tau}$};

    \draw[pile] (-0.8, -6.25) -- (0.3,-6.25) -- (0.3, -6.7);
    \draw[pile] (1.8, -6.25) -- (0.7,-6.25) -- (0.7, -6.7);

    \path[draw, fill=gray!10] (2,-6.8) -- (-1,-6.8) -- (-1.5,-8.7) -- (2.5,-8.7) -- (2,-6.8);
    \draw (0.5, -7.2) node {Invariance};
    \draw (0.5, -7.7) node {Decoder};
    
    \draw (0.5, -8.3) node {$p_{\theta}(\tilde{x}|z_c, z)$};
    \path[draw, fill=gray!40] (-1.5,-8.9) -- (2.5,-8.9) -- (2.5,-9.4) -- (-1.5,-9.4) -- (-1.5, -8.9);
    \draw (0.5, -9.15) node {$\pmb{\tilde{x}}$};
    
     \draw[pile] (6.8, -6.25) -- (5.7,-6.25) -- (5.7, -6.9);
     \draw[pile] (2.6, -9.1) -- (2.9, -9.1) -- (2.9, -7.5) -- (3.7, -7.5); 
    \path[draw, fill=gray!10] (7,-7.0) -- (4,-7.0) -- (3.5,-8.9) -- (7.5,-8.9) -- (7,-7.0);
    \draw (5.5, -7.6) node {Equivariant};
    \draw (5.5, -8.2) node {Reconstructor};
    
      \path[draw, fill=gray!40] (3.5,-9.1) -- (7.5,-9.1) -- (7.5,-9.6) -- (3.5,-9.6) -- (3.5, -9.1);
    \draw (5.5, -9.35) node {$\pmb{\hat{x}}$};

    \end{tikzpicture}
    }
    \caption{Structure of our basic MCE-VAE. The augmented encoder extracts the augmented latent variable $z_{aug}$ from the input data $x$, upon which the latent representations for variables $z_c, z,$ and $\tau$ are learned. The invariance decoder generates the canonical, invariant reconstruction $\tilde{x}$, which is passed together with $\tau$, to the equivariant reconstructor for obtaining the original input reconstruction $\hat{x}$.}
    \label{fig:model}
\end{figure}

We consider the information in the data as a combination of embeddings in multi-category finite invariant and equivariant spaces. In images of multi-category objects with varied spatial transformations, for example, the simplest partition can be the separation of categorical, variational, and transformational feature spaces. While the categorical distribution learns to distinguish different object categories(types), the variational distribution encodes the shape and shading variants within a category. As explained in the previous section, when the spatial transformations are confined on a Lie (differentiable) manifold, transformational distributions disentangle themselves from the categorical and variational ones by learning the Lie algebra of the transformation group under some canonical choice of the Lie group generators. In the VAE perspective, we learn the corresponding  latent subspace embedding for each of the factored distributions of the partitioned variables, denoted respectively by $z_c, z, \tau$. 
Figure \ref{fig:model} shows a simple example of this framework. 
The \textit{Augmented Encoder} generates an intermediate latent feature space $(z_{aug})$,  that feeds the extractor networks for learning each of  the separate embedding subspaces. The augmented latent variable $z_{aug}$ is learned from training the encoder, decoder, and  reconstructor networks on the dynamic data and the ELBO optimization derived in the following subsections. Similarly  $z_c, z, \tau$  and their respective embedding spaces can be subsequently extracted and learned from $z_{aug}$ using trained extractor and reconstructor networks. Estimating the latent space utilizes the reparameterization trick, $   (z_c, z, \tau) = f\left(\mathrm{EncoderNN}_\phi(x), \epsilon \right),$ where $\phi$ represents the joint set of learnable parameters of the encoder and extractor networks, $\epsilon \sim \mathcal{N}\left(0, \mathbb{I}_d \right)$ represent a sample drawn from a $d = |z_c| + |z| + |\tau|$ dimensional multivariate normal distribution, and $f:\mathbb{R}^{2d} \times \mathbb{R}^d \to \mathbb{R}^d$ denotes the deterministic sampling function. 

The first component of the decoder units consists of an invariance decoder. Independent of the spatial transformational latent space, the invariance decoder learns a canonical reconstruction $(\tilde{x})$ that is invariant under Lie group transformations. The second component is the equivariant reconstructor, a spatial transformation network that learns the exponential map of the Lie Algebra and applies it on the canonical reconstruction $(\tilde{x})$ to obtain an equivariant reconstructed representation of the data.
In an invariance-enforced setting, if $p_\theta(\tilde{x}|x)$ denotes the conditional distribution of the canonical reconstruction $(\tilde{x})$ for some given input data $(x)$, it is invariant under transformations on $x$, i.e. for any $M \in L$,
\begin{equation}
p_\theta(\tilde{x}|x) = p_\theta(\tilde{x}|Mx) 
    \label{eqn:invariance-condn}
\end{equation}

\subsection{Formulation and Variational Lower bounds} \label{sec:elbo}
The invariance-informed objective of the proposed VAE is to maximize the marginal log-likelihood of $\log p_\theta(x)$ while subject to the condition in Eq. \ref{eqn:invariance-condn}. Following KKT criterion, the objective function can be restated as,
\begin{equation}
%$
    \mathcal{L} = \log p_\theta(x) - \alpha\mathbb{E}_{\rho(M)}\mathbf{D}(p_\theta(\tilde{x}|x), p_\theta(\tilde{x}|Mx))
    \label{eqn:objective}
%$
\end{equation}
Here $\alpha > 0$ is a tunable hyperparameter set as 1.0 for the rest of the paper. $\mathbf{D}(\cdot,\cdot)$ is a non-negative divergence metric between two distributions satisfying $\mathbf{D}(p,p) \le \mathbf{D}(p,q)$ for all distributions $p, q$ defined on the same manifold. 
$\rho(M)$ is a distribution defined on the members of the Lie group $L$. In this paper we sample $M$ from a uniform distribution on a finite support for the coefficient vector $\tau$ for unsupervised learning. The domain of the finite support can either be determined by the nature of the transformation (e.g. in case of rotation) or from the finite dimensions of the feature space (e.g. the physical dimensions of image data).  For a supervised training network, where both $x$ and its pre-transformed ground truth $x_{gt}$ are known, $\rho(M) = \delta(M - M^*)$ where $x_{gt} = M^*x$.

The first part of the objective function in Eq. \ref{eqn:objective} can be constrained by an Evidence Lower Bound (ELBO) using the standard VAE algebra.
We start with the basic version of MCE-VAE disentangling semantic and transformation latent representations with only one $z$ and one $\tau$. The variational lower bound can be written as
\begin{equation}
    \begin{aligned}
    ELBO & =   \mathbb{E}_{q_\phi}\log\frac{p_\theta(z_c, z, \tau, x)}{q_\phi(z_c, z, \tau|x)}\\
    & = \mathbb{E}_{q_\phi}\left[\log p_\theta(x|z_c, z, \tau) \right] 
    - \mathrm{KL}(q_\phi(z_c|x) || p_\theta(z_c) ) \\
    & \quad - \mathrm{KL}(q_\phi(z|x) || p_\theta(z) )
    - \mathrm{KL}(q_\phi(\tau|x) || p_\theta(\tau) )
    \end{aligned}
\end{equation}
where the first term is the overall reconstruction likelihood, and the following terms represent the KL-divergence of the posterior for different components of the latent space.

\subsection{Invariance-informed Learning}
In case of supervised learning where we choose the distribution on $M$ to be a delta function, the divergence term in Eq. \ref{eqn:objective} becomes,
%\begin{equation}
$
    \mathbb{E}_{\rho(M)}\mathbf{D}(p_\theta(\tilde{x}|x), p_\theta(\tilde{x}|Mx)) = \mathbf{D}(p_\theta(\tilde{x}|x), p_\theta(\tilde{x}|x_{gt}))
    \label{eqn:div-sup}
$
%\end{equation}
which can be faithfully represented by the negative binary cross-entropy between reconstructed $\tilde{x}$ and ground truth $x_{gt}$.
However, in an unsupervised setting, the distribution $p_\theta(\tilde{x}|x)$ is an intractable one. However, intuitively we can construct the divergence metric by requiring that both the canonical reconstruction $(\tilde{x})$ and the latent space representation $(z_c, z)$ are invariant under group transformations, which can be quantitatively enforced as
\begin{align}
\mathbf{D}(p_\theta(\tilde{x}|x), p_\theta(\tilde{x}|Mx)) \approx & \left(z_c(x) - z_c^*(Mx)\right)^2 \nonumber \\
& + \left(z(x) - z^*(Mx)\right)^2  \nonumber \\
& - \mathrm{BCE}\left(\tilde{x}(x), \tilde{x}^*(Mx)\right)
\end{align}
where $\mathrm{BCE}(\cdot , \cdot)$ represents binary cross-entropy. The variables marked with an asterick represents they are evaluated at the current value of the NN parameters $(\theta, \phi)$ and don't contribute to backpropagation of gradients.

\subsection{Cluster Extractor}
Now we emphasize the cluster learning part of our proposed network. We assume that the latent space representation of different categories are represented as a probability mixture model. Hence, the $z_c$ variables are set with the following prior and posterior distributions,
\begin{align}
    p_\theta(z_c) &= \sum_{i=1}^{n_C} p_i p_{\theta,i}(z_c) \label{eqn:cat-prior} \\
    q_\phi(z_c|x) &= \sum_{i=1}^{n_C}\pi_i q_{\phi,i}(z_c|x)
    \label{eqn:cat-post}
\end{align}
where $n_C$ is the number of mixture components, $p_i$ and $\pi_i$ respectively represent the prior and posterior marginal distributions of the cluster components with $\sum_i p_i = \sum_i \pi_i = 1$. 
% & = \frac{1}{n_C}\sum_{i=1}^{n_C}\mathcal{N}\left(\mu_i, \mathbb{I}_{n_C} \right)

%where $n_C = |z_c|$ and  $\mu_i = (\delta_{ij})_{j=1}^{n_C}$. 
%The posterior is also assumed to be an equimixture Gaussian,

% \begin{align}
%      \nonumber \\ 
%     & = \frac{1}{n_C}\sum_{i=1}^{n_C}\frac{1}{\sqrt{2\pi\sigma(x)^2}}\exp\left[-\frac{z_{c,i} - \mu_i(x)}{2\sigma(x)^2}\right] \times \mathcal{N}\left(0, \mathbb{I}_{n_C-1}  \right)
    
% \end{align}
Using Eqns. \ref{eqn:cat-prior} and \ref{eqn:cat-post}, we can obtain an upper bound on the expression for KL divergence on $z_c$. Let's start with the expression for KL divergence,
%\begin{equation}
$
    \mathrm{KL}\left( q_\phi(z_c|x) || p_\theta(z_c) \right) = \mathbb{E}_{q_\phi}\left[ \log q_\phi(z_c|x) -  \log p_\theta(z_c) \right]
$
%\end{equation}
First, using concavity of the log function, the second part in the right side can be bounded as
\begin{align}
    \mathbb{E}_{q_\phi}\log\left(p_\theta(z_c) \right) & =\mathbb{E}_{q_\phi}\log\left(\sum_{k}p_ip_{\theta,k}(z_c) \right) \nonumber \\
    & \ge \sum_{k} p_k \mathbb{E}_{q_{\phi,k}} \log p_{\theta,k}(z_c) 
    \label{eqn:bound-p}
\end{align}
For the other part,
\begin{align}
    & \mathbb{E}_{q_\phi}\log\left(q_\phi(z_c|x) \right) \nonumber \\
    =& \int q_\phi(z_c|x) \log\left(q_\phi(z_c|x)\right) dz_c \nonumber \\
    =& \int \left[\sum_{k} \pi_k q_{\phi,k}(z_c|x)\right] \log\left(\sum_{k'} \pi_{k'} q_{\phi,k'}(z_{c}|x)\right) dz_c \nonumber \\
     \le & \int \sum_{k} \pi_k q_{\phi,k}(z_c|x) \log\left(q_{\phi,k}(z_{c}|x)\right) dz_c \nonumber\\
     = & \sum_{k} \pi_k \mathbb{E}_{q_{\phi,k}}\log q_{\phi,k}(z_{c}|x)
    \label{eqn:bound-q}
\end{align}
where the inequality follows from log-sum inequality. Together with  Eqns. \ref{eqn:bound-p} and \ref{eqn:bound-q}, the KL divergence can be bounded from above as 
\begin{align}
    \mathrm{KL}\left( q_\phi(z_c|x) || p_\theta(z_c) \right) \le & \sum_{k} \pi_k \mathbb{E}_{q_{\phi,k}}\log q_{\phi,k}(z_{c}|x) \nonumber \\
    & - \sum_{k} p_k \mathbb{E}_{q_{\phi,k}} \log(p_{\theta,k}(z_c) ) \label{cluster-KL}
\end{align}
In the special case of $\pi_k = p_k$, Eqn.\ref{cluster-KL} reduces to,
\begin{equation}
    \mathrm{KL}\left( q_\phi(z_c|x) || p_\theta(z_c) \right) \le \sum_{k} \pi_k \mathrm{KL}\left( q_{\phi,k}(z_c|x) || p_{\theta,k}(z_c) \right)
\end{equation}
\section{Experimental Results}\label{sec:exp}
\begin{figure*}
\centering
 %\vspace{-2mm}
  \begin{subfigure}[b]{0.75\textwidth} 
  %\subfloat[Supervised SE(2)]{
     \centering
     \includegraphics[width=\textwidth, trim=1.5in 0.4in 1.5in 0.1in, clip]
     {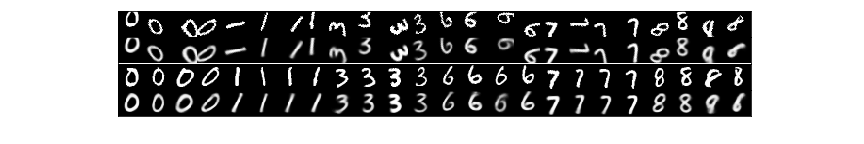}
     \caption{Supervised SE(2)}
     \label{fig:sup_SE2}
   %  }
  \end{subfigure}
   
  \begin{subfigure}[b]{0.75\textwidth}
  %\subfloat[Unsupervised SE(2)]{
     \centering
     \includegraphics[width=\textwidth, trim=1.5in 0.4in 1.5in 0.1in, clip]
     {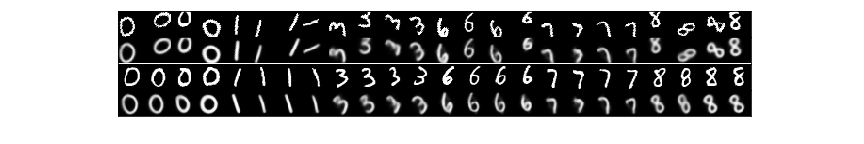}
     \caption{Unsupervised SE(2)}
     \label{fig:unsup_SE2}
   %  }
  \end{subfigure}
  
  %\vspace{-2mm}
  \begin{subfigure}[b]{0.75\textwidth} 
  %\subfloat[Supervised SO(2)]{
     \centering
     \includegraphics[width=\textwidth, trim=1.5in 0.4in 1.5in 0.1in, clip]
     {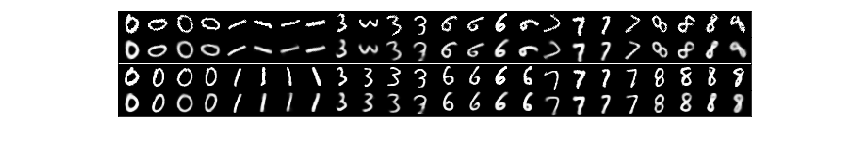}
     \caption{Supervised SO(2)}
     \label{fig:sup_SO2}
    % }
  \end{subfigure}
  
  %\vspace{-2mm}
  \begin{subfigure}[b]{0.75\textwidth}
  %\subfloat[Unsupervised SO(2)]{
     \centering
     \includegraphics[width=\textwidth, trim=1.5in 0.4in 1.5in 0.1in, clip]
     {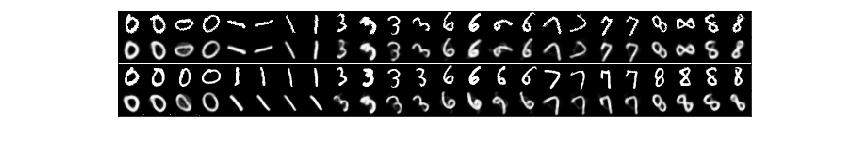}
     \caption{Unsupervised SO(2)}
   %  }
     \label{fig:unsup_SO2}
  \end{subfigure}
  %\hspace{-8mm}
\caption{Examples of dataset from supervised and unsupervised training on SO(2) and SE(2) transformed datasets. In each image the top row represents the transformed imaged $(x)$ that was the input to the network. The second row represents the reconstructed equivariant image $(\hat{x})$. The third row represents the ground truth $(x_{gt})$ which is only used for the purpose of training in supervised settings. Finally, the fourth row is the canonical invariant representation $(\tilde{x})$ learned by the network}
\label{fig:all_reco}
\end{figure*}

To test the effectiveness of our proposed model, we examined the reconstruction accuracy of our model on MNIST images of handwritten digits with rigid spatial transformations. The dataset has been created by performing arbitrary rotations and translations on the MNIST images where the rotation angles are limited between $\pm 90^o$ and the translations are confined within $\pm 25\%$ of the image dimension. The 70000 MNIST images are split into 6:1 ratio as training and validation datasets. 
The augmented encoder is composed of a four-layer 2D convolutional network with batch normalization and ReLU activation. 
The cluster and variational extractor units are implemented as MLPs with two hidden layers with sigmoid activation and 512 nodes per hidden dimension. On the other hand, the transformation extractor with a similar structure has a 32 nodes per hidden dimension. The cluster extractor gives output in the latent dimension of $10$ and the variation extractor is implemented in a latent dimension of 3. 
We have chosen a Gaussian Mixture Model for the clustering latent variables.
\begin{align}
p_{\theta,i} (z_c) & = \mathcal{N}\left(\mu_i, \mathbb{I}_{n_C} \right) \label{eqn:gaus-prior}\\
q_{\phi,i}(z_c|x) & = \frac{1}{\sqrt{2\pi\sigma_i(x)^2}}\exp\left[-\frac{\left(z_{c,i} - \mu_i(x)\right)^2}{2\sigma_i(x)^2}\right] \nonumber \\
&\times \mathcal{N}\left(0, \mathbb{I}_{n_C-1}  \right) \label{eqn:gaus-post}
\end{align}
where we have chosen  $n_C = |z_c|$ and  $\mu_i = (\delta_{ij})_{j=1}^{|z_C|}$. For simplicity, we also set both the prior and posterior as equimixture Gaussians with $p_c = \pi_c = \frac{1}{n_C}$. To ensure back-propagation across the cluster extractor, we use the reparameterization trick-
\begin{equation}
    \begin{aligned}
    \mu(x), \log\sigma(x) &= \mathrm{EncoderNN}_{(\Phi, \phi_c)} \left(x\right) \\
    z_{c_i} &= \mu_i(x) + \epsilon_i\sigma_i(x)
    \end{aligned}
\end{equation}
     
%\begin{wrapfigure}{r}{0.5\textwidth}
\begin{figure*}
    \begin{subfigure}[b]{0.6\textwidth}
       \centering
    \includegraphics[width=0.9\textwidth]{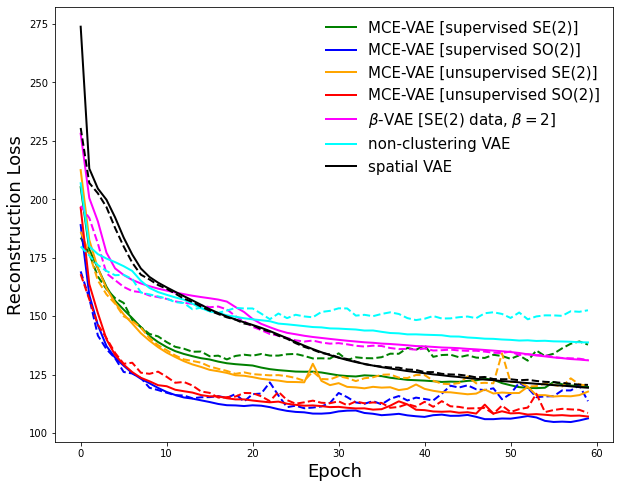}
    \caption{\label{fig:loss}}
    \end{subfigure}%
    \begin{subfigure}[b]{0.4\textwidth}
    \centering
    \includegraphics[width=0.9\textwidth, trim=0.5in 0.5in 0.2in 0.2in, clip]{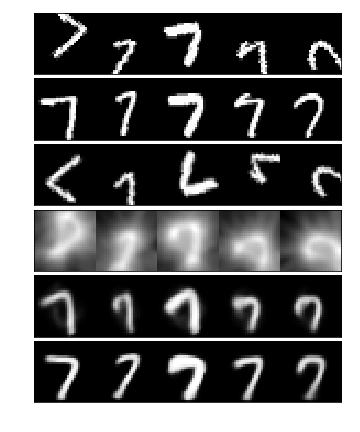}
    \caption{\label{fig:reco_comp}}
    \end{subfigure}
    \caption{
    (a) Training (solid) and test (dotted) reconstruction loss of our model with both supervised training and unsupervised training. The result is compared with spatial-VAE and $\beta$-VAE ($\beta = 2$). We additionally compare the results of our model with a non-clustering prior and posterior on the $z_c$ latent space
    (b) Comparison of canonical reconstruction of random SE(2) transformed images from different models. The top row represents the input data and the second row is the ground truth. Third and fourth rows represent reconstructions from Affine-VAE and Spatial-VAEs. The last two rows are obtained from unsupervised and supervised MCE-VAE networks 
    }
\end{figure*}
    
%\end{wrapfigure}     
     
Finally, the invariance decoder is implemented as a multi-layer GatedDense network with 300 nodes per hidden layer. Limiting our explorations to rigid transformations in 2D, the equivariant reconstructor is implemented as a spatial transformation network.

% \begin{wrapfigure}{r}{0.5\textwidth}
%     \centering
%     \includegraphics[width=0.5\textwidth, trim=0.5in 0.5in 0.2in 0.2in, clip]{reco_comp.png}
%     \caption{\label{fig:reco_comp} Comparison of canonical reconstruction of random SE(2) transformed images from different models. The top row represents the input data and the second row is the ground truth. Third and fourth rows represent reconstructions from Affine-VAE and Spatial-VAEs. The last two rows are obtained from our  unsupervised and supervised MCE-VAE networks 
%     }
% \end{wrapfigure}
% In order to test the performance of our model in terms of reconstruction accuracy, we trained MCE-VAE on a variant of MNIST with spatial transformation and measured the reconstruction loss. For this task, we generated variants of MNIST with spatial transformations. In MNIST-SO2 (\cref{fig:SO2}), each MNIST digit is rotated with a random rotation angle $r\pi$, where $r \sim \mathcal{N}(0, 0.5)$. Similarly, in MNIST-SE2 (\cref{fig:SE2}), each MNIST digit is rotated with a random rotation angle $r_a\pi$ and translated by $r_xw_x$ horizontally and $r_yw_y$ pixels vertically, where $r_a, r_x, r_y \sim \mathcal{N}(0, 0.5)$ and $w_x, w_y$ denotes the horizontal and vertical image width ($28$ for MNIST digits). 
 Each of the networks has been trained for 60 epochs with minibatches of size 100 using the Adam optimizer with a learning rate of $10^{-3}$. In Figure \ref{fig:all_reco}, we show the equivariant as well as the canonical, invariant reconstruction for supervised and unsupervised training on SO(2) and SE(2) transformed datasets. All the images in this figure are taken from the validation dataset. For the same group transformation, the reconstruction quality is  similar for both training modes, which can also be quantitatively verified from Figure \ref{fig:loss}- the reconstruction accuracy of our model with unsupervised training is converging to a very similar level of the supervised accuracy. In the same plot we compare the reconstruction loss  with benchmark models- spatial-VAE, affine-VAE, and $\beta$-VAE on SE(2) transformed MNIST. Both the supervised and unsupervised version of our model outperform the benchmarks in terms of both accuracy and learning rate. 

In Figure \ref{fig:reco_comp} we visually compare the quality of invariant reconstruction from our proposed architecture with the ones obtained from benchmark models.  
%against the Ideally, the invariant semantic representation should be the same for the same object with different orientations. However, as shown in \cref{fig:metric} (left), 
As illustrated, methods like spatial-VAE and affine-VAE fail to guarantee invariance while the canonical reconstruction has almost identical orientation independent of the transformation that obtains the input data.

\begin{figure*}[b]
\centering
  \begin{subfigure}[b]{0.46875\textwidth} 
     \centering
     \includegraphics[width=\textwidth, trim=1in 0.7in 1in 0.7in, clip]
     {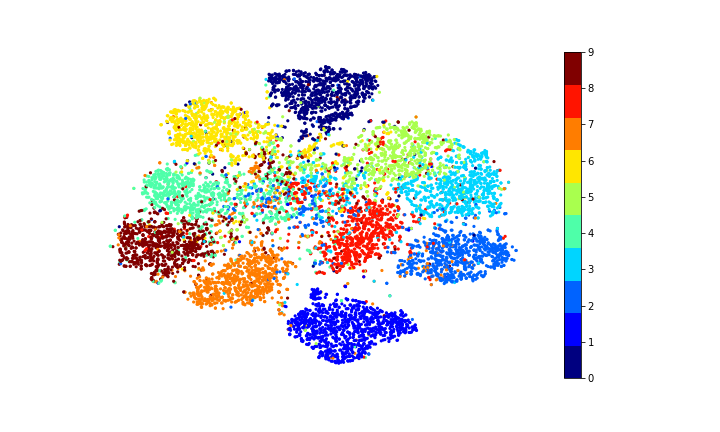}
     \caption{Supervised SE(2)}
     \label{fig:sup_SE2}
  \end{subfigure}%
  \begin{subfigure}[b]{0.46875\textwidth}
     \centering
     \includegraphics[width=\textwidth, trim=1in 0.7in 1in 0.7in, clip]
     {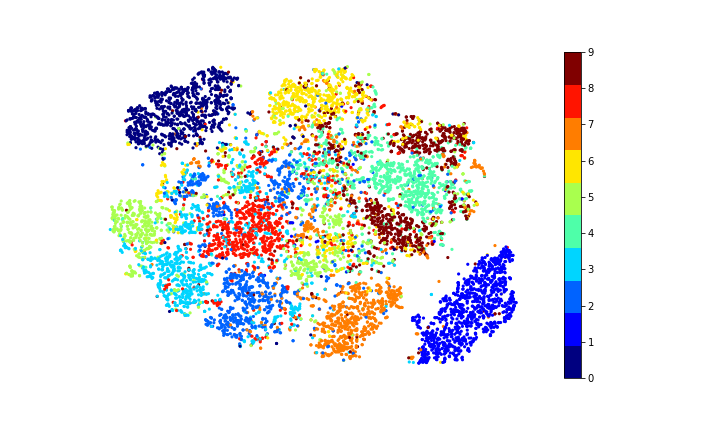}
     \caption{Unsupervised SE(2)}
     \label{fig:unsup_SE2}
  \end{subfigure}
  
  \begin{subfigure}[b]{0.46875\textwidth} 
     \centering
     \includegraphics[width=\textwidth, trim=1in 0.7in 1in 0.7in, clip]
     {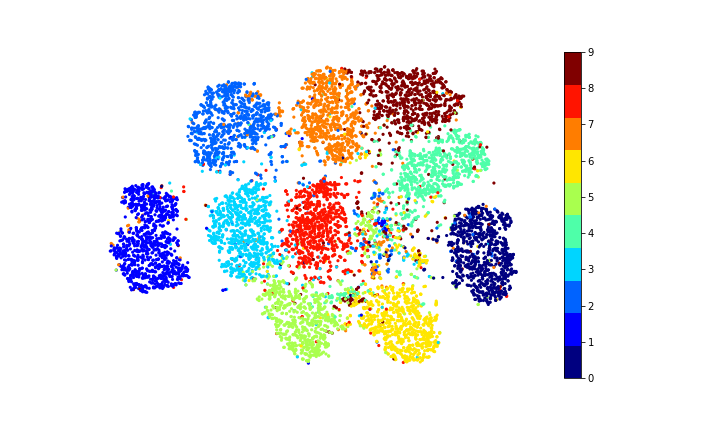}
     \caption{Supervised SO(2)}
     \label{fig:sup_SO2}
  \end{subfigure}%
  \begin{subfigure}[b]{0.46875\textwidth}
     \centering
     \includegraphics[width=\textwidth, trim=1in 0.7in 1in 0.7in, clip]
     {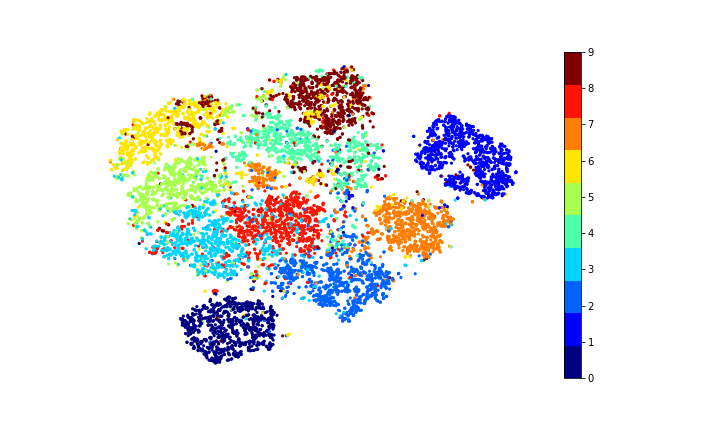}
     \caption{Unsupervised SO(2)}
     \label{fig:unsup_SO2}
  \end{subfigure}
  
  \begin{subfigure}[b]{0.46875\textwidth} 
     \centering
     \includegraphics[width=\textwidth, trim=1in 0.7in 1in 0.7in, clip]
     {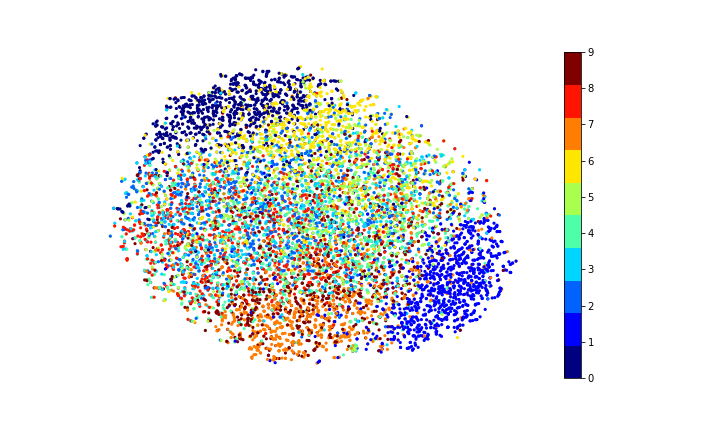}
     \caption{Supervised non-clustering SE(2)}
     \label{fig:sup_SE2_nocluster}
  \end{subfigure}%
  \begin{subfigure}[b]{0.46875\textwidth}
     \centering
     \includegraphics[width=\textwidth, trim=1in 0.7in 1in 0.7in, clip]
     {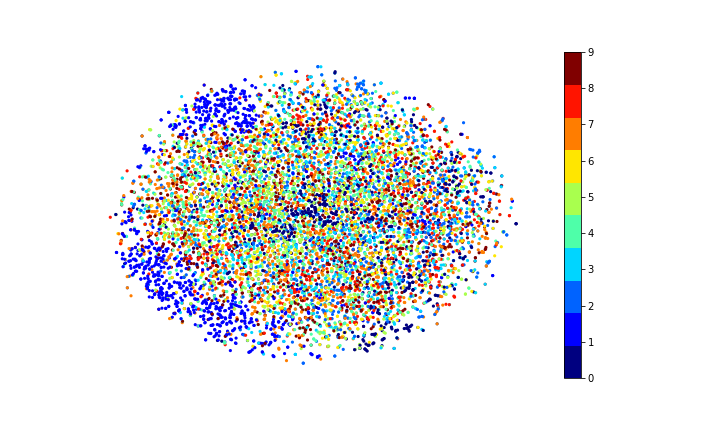}
     \caption{SE(2) with $\beta$-VAE ($\beta = 2.0$)}
     \label{fig:beta_VAE}
  \end{subfigure}
\caption{Visual representation of clustering according to digit types in MNIST images for different training settings obtained from a 2D TSNE analysis on latent feature space. Figures (a)--(d) are obtained from the $z_c$ latent subspace of an MCE-VAE network with a GMM posterior. In Figure~(e), a non-clustering, single, multivariate Gaussian posterior is used for the $z_c$ subspace with rigid body transformations embedded on the Lie manifold. For Figure (f), a $\beta$-VAE is trained using a 16 dimensional latent space by turning off the multi-cluster and the equivariance extractors with a choice of $\beta = 2$.
}
\label{fig:all_cluster}
\end{figure*}

A major improvement of this model is its capacity to learn clustering in the latent variable space. We show the clustering capacity of the proposed network in Figure \ref{fig:all_cluster} by performing a TSNE analysis \cite{van:2008} on the $z_c$ latent space. We  can clearly see efficient clustering according to digit types in both supervised and unsupervised settings for SO(2) and SE(2) transformed MNIST images. Using the GMM prior and posterior has the clear benefit of superior cluster identification over a single, multivariate, diagonal Gaussian posterior even when the model is trained to learn the rigid body transformation over a Lie manifold (Figure \ref{fig:sup_SE2_nocluster}). Our model also performs significantly better than usual disentangling networks like the $\beta$-VAE (Figure \ref{fig:beta_VAE}) in cluster learning. 

\section{Conclusion}
This paper presents a generalized unsupervised variational auto-encoder multi-network model,  MCE-VAE, that simultaneously learns to cluster image data while learning its canonical representation as well as a Lie manifold constrained of image. The latent space representations emphasizes the forced learning of invariance to ensure the disentanglement of the multi-clusters of categorical, variational, and spatial transformation representations. In doing so, this model addresses a relatively unexplored genre of unsupervised learning in computer vision- group invariant clustering in an equivariant setting. Through visualizations of the learned latent invariant representations and through the quantitative measurements, we demonstrate that MCE-VAE outperforms the state-of-the-art equivariant learning  approaches for VAE. Our MCE-VAE has the potential to simultaneously learn multiple invariant (categorical and semantic) and equivariant Lie group representations. Moreover, MCE-VAE can be extended to other continuous Lie group transformations learning of multiple objects. On the application side, we believe this work can lead to a new approach in solving tasks like multi-object tracking and  trajectory recognition in noisy dynamic scenes.

%\begin{acknowledgements} % will be removed in pdf for initial submission,
                         % so you can already fill it to test with the
                         % ‘accepted’ class option
%    Briefly acknowledge people and organizations here.

 \emph{Acknowledgement:} 
 %acknowledgements go in this section.
 This research was supported in part by a grant from NIH - R01GM117594, by the Peter O’Donnell Foundation and in part from a grant from the Army Research Office accomplished under Cooperative Agreement Number W911NF-19-2-0333. The work of A.R. is supported by U.S. Department of Energy, Office of High
Energy Physics under Grant No. DE-SC0007890 The views and conclusions contained in this document are those of the authors and should not be interpreted as representing the official policies, either expressed or implied, of the Army Research Office or the U.S. Government. The U.S. Government is authorized to reproduce and distribute reprints for Government purposes notwithstanding any copyright notation herein.
%\end{acknowledgements}

\bibliography{MultiClusterVAE}
\bibliographystyle{plainnat}

\newpage

\end{document}